# Chest X-Ray Analysis of Tuberculosis by Deep Learning with Segmentation and Augmentation

Sergii Stirenko*, Yuriy Kochura, Oleg Alienin, Oleksandr Rokovyi, and Yuri Gordienko
National Technical University of Ukraine
"Igor Sikorsky Kyiv Polytechnic Institute",
Kyiv, Ukraine
*sergii.stirenko@gmail.com

Peng Gang** and Wei Zeng

Huizhou University,
Huizhou City, China

**peng@hzu.edu.cn

*Abstract*—The results of chest X-ray (CXR) analysis of 2D images to get the statistically reliable predictions (availability of tuberculosis) by computer-aided diagnosis (CADx) on the basis of deep learning are presented. They demonstrate the efficiency of lung segmentation, lossless and lossy data augmentation for CADx of tuberculosis by deep convolutional neural network (CNN) applied to the small and not well-balanced dataset even. CNN demonstrates ability to train (despite overfitting) on the pre-processed dataset obtained after lung segmentation in contrast to the original not-segmented dataset. Lossless data augmentation of the segmented dataset leads to the lowest validation loss (without overfitting) and nearly the same accuracy (within the limits of standard deviation) in comparison to the original and other pre-processed datasets after lossy data augmentation. The additional limited lossy data augmentation results in the lower validation loss, but with a decrease of the validation accuracy. In conclusion, besides the more complex deep CNNs and bigger datasets, the better progress of CADx for the small and not well-balanced datasets even could be obtained by better segmentation, data augmentation, dataset stratification, and exclusion of non-evident outliers.

## I. Introduction

Due to relatively cheap price and easy accessibility chest X-ray (CXR) imaging is used widely for health monitoring and diagnosis of many lung diseases (pneumonia, tuberculosis, cancer, etc.). Manual analysis and detection by CXR of marks of these diseases is carried out by expert radiologists, which is a long and complicated process. Nevertheless, the modern evolution of general-purpose graphic processing cards (GPU) hardware [1] and software for medical image analysis [2], especially deep learning techniques [3], allows scientists to detect automatically many lung diseases from CXR images at a level exceeding certified radiologists [4]. Despite these successes the strong belief exists among experts that deep learning techniques become efficient for the very big datasets ($>10^4$ images), because for the smaller datasets ($<10^3$ images) they produce bad predictions (if any at all) with the very low accuracies. This paper is dedicated to the description of the lung segmentation technique in combination with lossless and lossy data augmentation which allow us to get the statistically reliable predictions of lung diseases (availability of tuberculosis) for such a small dataset ($<10^3$ images) even.

## II. Problem and Related Work

Recently, promising results were obtained in the field of computer-aided medical image analysis for an assessment of lung diseases by deep learning from CXR image analysis [2,4-6]. Availability of open datasets with labeled CXR images allowed data scientists to apply their deep learning algorithms for anatomical structure detection, segmentation, computer-aided detection (CADe) of suspicious regions, and computer-aided diagnosis (CADx) [2]. Various datasets with CXR images were released for public domain recently: JSRT dataset with 247 images of cancer [7]; LIDC dataset with ~$10^3$ images [8]; Montgomery County (MC) dataset with 138 images, Shenzhen Hospital (SH) dataset with 662 images [9,10] (Fig. 1), and ChestX-ray14 [11], which is the largest publicly available database of CXR images to the moment (112,120 images of 30,805 unique patients). Several hardships with CADx of lung diseases on the basis of the small datasets ($<10^3$ images) are related with presence of high variability of image sizes, image quality, and influence of other regions outside of lungs (heart, shoulders, ribs, clavicles, etc.). This variability is originated by presence of patients of different gender, nationality, age, substance abuse, professional activity, general health state, and other parameters of patients under investigations. Such variability is strengthened by some additional artifacts like labels created by radiologists and/or doctors (like white rectangles with text and symbols shown in top part on Fig.1a).

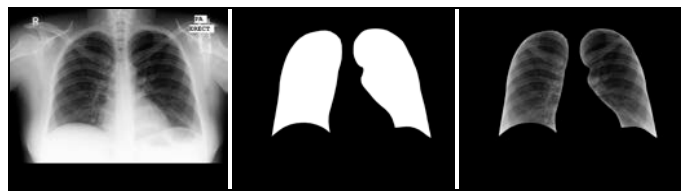

Fig. 1. Examples of the original image from SH dataset (left), its lung mask (central), and the segmented result after application of the mask (right).

Recently, efficiency of some dimensionality reduction techniques, like lung segmentation, bone shadow exclusion, and t-distributed stochastic neighbor embedding (t-SNE) for exclusion of outliers, was demonstrated for analysis of CXR

2D images to identify marks of lung cancer [5-6]. Despite these attempts, CADx remains a challenge in medical image applications, especially, in the view of high variability of lungs and other regions outside of lungs due to different personal parameters of patients under investigations. The aim of this work is to demonstrate the efficiency of the proposed lung segmentation technique in combination with lossless and lossy data augmentation methods to get the statistically reliable predictions of lung diseases (availability of tuberculosis) even for the relatively small datasets like SH ($<10^3$ images) [9,10].

### III. EQUIPMENT, METHODS, AND DATA

The deep CNN [5] was trained on various general-purpose graphic processing cards (GPU) by NVIDIA: Tesla K40 (Kepler microarchitecture), GTX 1080 Ti (Pascal), and the latest and most powerful GPU-card, Tesla V100 (Volta).

TABLE I. PERFORMANCE COMPARISON

|  | Performance (single), TFLOPS | Bandwidth, GB/s | Cores | Training Time (hh:mm) | Speedup |
|---|---|---|---|---|---|
| K40 | 4.3-5.0 | 288 | 2880 | 12:08 | 1 |
| GTX 1080 Ti | 10.6 | 484 | 3584 | 00:50 | 14.56 |
| V100 | 14.0 | 900 | 5120 | 00:33 | 22 |

The official data on performance, bandwidth, number of cores, and the actual training times with the correspondent speedups (for the same dataset and runtime configuration) are shown in Table 1. It worth to note that the CNN model was not optimized here for calculations by tensor processing unit (TPU) in V100 card. That is why its speedup is not so impressive in this work, but the results on the optimized version will be reported elsewhere [12].

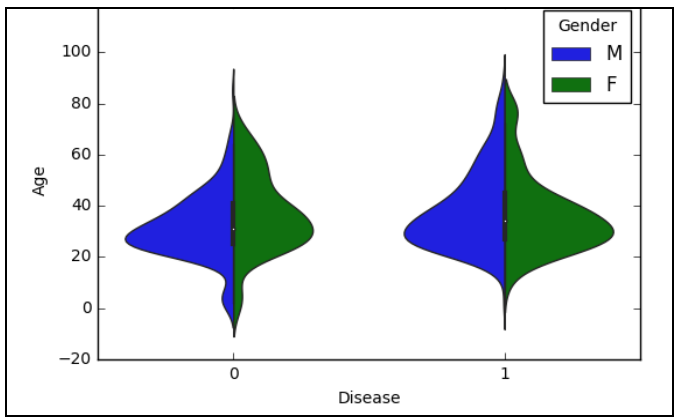

Fig. 2. Combined distribution of genders and ages among images without (0) and with (1) disease marks.

Shenzhen Hospital (SH) dataset of CXR images was acquired from Shenzhen No. 3 People's Hospital in Shenzhen, China. It contains normal and abnormal CXR images with marks of tuberculosis. The exploratory data analysis shown that this small dataset is a not well-balanced dataset with regard to availability/absence of disease, age, gender (Fig. 2). The image widths and heights are not equal and these differences lead to the wide and asymmetric distribution of image areas (Fig. 3). The current and previous attempts to perform training for such small CXR datasets without any pre-processing were performed and failed [5,6]. Fortunately, "external segmentation" of the left and right lung fields in CXR images (exclusion of outside regions which are not pertinent to lungs) was demonstrated to be effective to provide successful training and, moreover, to increase the accuracy of predictions [5,6].

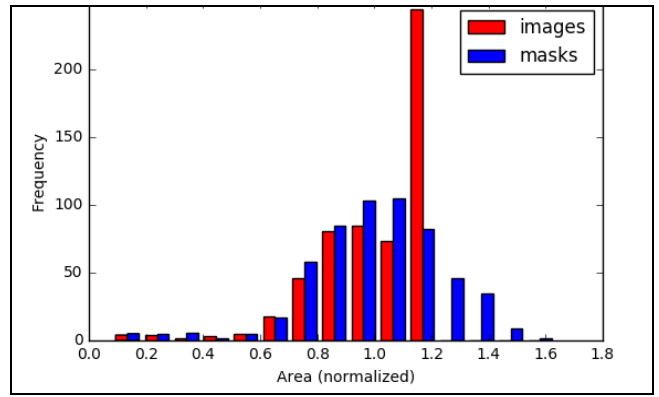

Fig. 3. Distributions of image (Fig. 1, left) and mask (Fig. 1, central) areas.

To perform lung segmentation, i.e. to cut the left and right lung fields from the lung parts in standard CXRs, the manually prepared masks were used (Fig. 1, central) that allowed to get the additional dataset from SH, namely, the segmented SH dataset (Fig. 1, right). Despite the high variability of lung segmentation masks (due to the different lung shapes and borders) the distribution of lung mask areas was much closer to the normal one in comparison to the distribution of image areas, which was asymmetric and contained the long tail for smaller sizes (Fig. 3). Some examples of the most similar, dissimilar, and average lung masks are shown in Fig. 4. This variability is explained by presence of outliers, for example, some portion of children with the smaller lungs (see the local peaks for age<20 in Fig. 2).

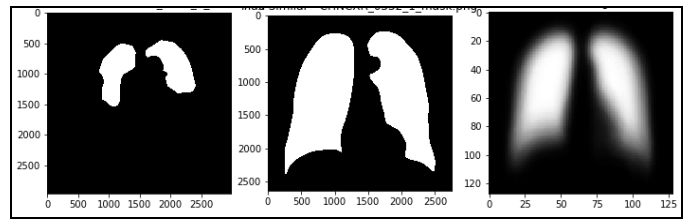

Fig. 4. The most dissimilar (left), the most similar (central), and the average (right) lung masks.

### IV. RESULTS

The deep CNN [5] was trained in GPU mode by means of TensorFlow machine learning framework [13] to predict presence (336 images) or absence (326 images) of tuberculosis in them. The dataset was split into 8:1:1 parts for training, validation and test parts respectively. Images were rescaled 1./255 and resized to 2048×2048 and distributed among training, validation and test parts. Several training and validation runs up to the 100 epochs for the CNN with dropout on CXR images from the segmented SH dataset were performed for: a) the segmented SH dataset; b) the segmented SH dataset with lossless and c) lossy data augmentation.

## A. Segmented Dataset

The examples of accuracy and loss values after some runs are shown in Fig. 5 (accuracy) and Fig. 6 (loss): for training and validation, after averaging their values over the runs (red line in Fig. 5 and Fig. 6), and after smoothing the averaged time series (blue line in Fig. 5 and Fig. 6) by the locally weighted polynomial regression method [14].

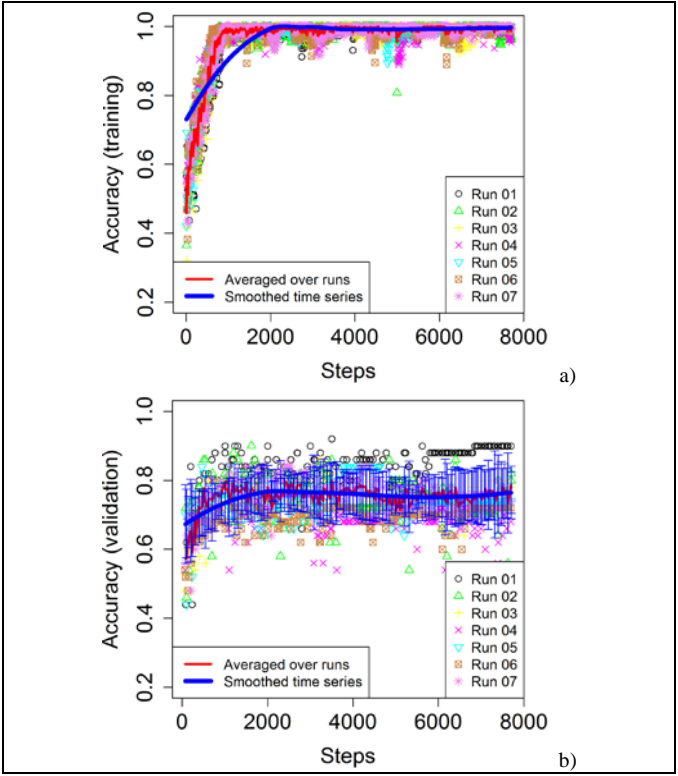

Fig. 5. Accuracy during some cross-validation runs for the segmented CXR images from SH dataset: training (a) and validation (b).

The pronounced overtraining (overfitting) can be observed after comparison of these training and validation results (Fig. 5 and Fig. 6). The averaged and smoothed value of training accuracy (Fig. 5a) is going with steps up to the theoretical maximum of 1 and the training loss (Fig. 6a) is going to 0. At the same time the averaged and smoothed value of the validation accuracy (Fig. 5b) is no more than 0.76 and the validation loss (Fig. 6b) is very high and growing. Such overfitting can be explained by the small size of the dataset, which can be decreased by the well-known data augmentation methods. For this purpose, the lossless and lossy types of data augmentation were applied and their role investigated.

## B. Effect of Lossless Data Augmentation

The lossless data augmentation for 2D images included the following transformations: mirror-like reflections (left-right and up-down) and rotations by $90n$ degrees, where $n = 1,2,3$. This allowed to increase efficiently (by 8 times) the whole dataset and obtain the more realistic results on accuracy and loss during training and validation. The training accuracy did not exceed 0.9 (Fig. 7a), the validation accuracy did not exceed 0.7 (Fig. 7b), and the training loss (Fig. 8a) was going to 0 with much lower rate.

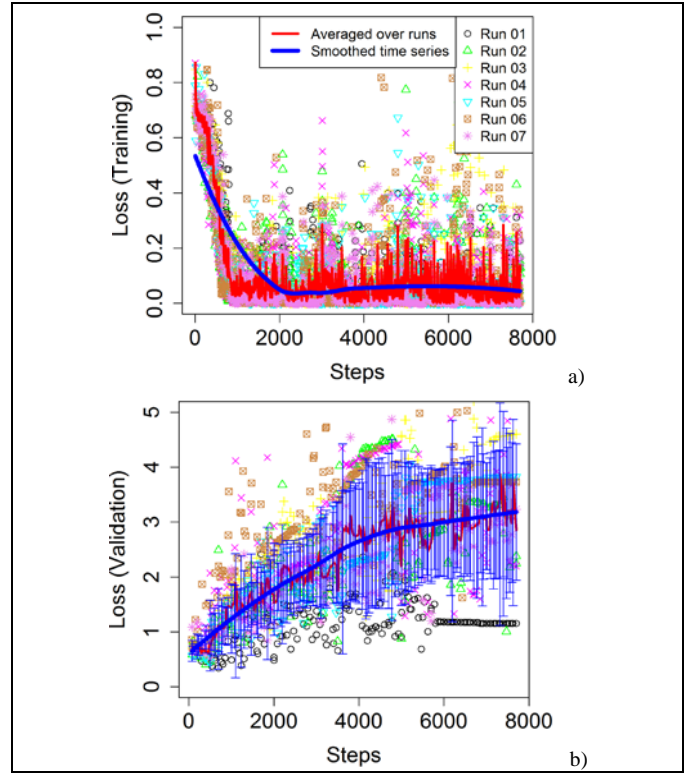

Fig. 6. Loss during some cross-validation runs for the segmented CXR images from SH dataset: training (a) and validation (b).

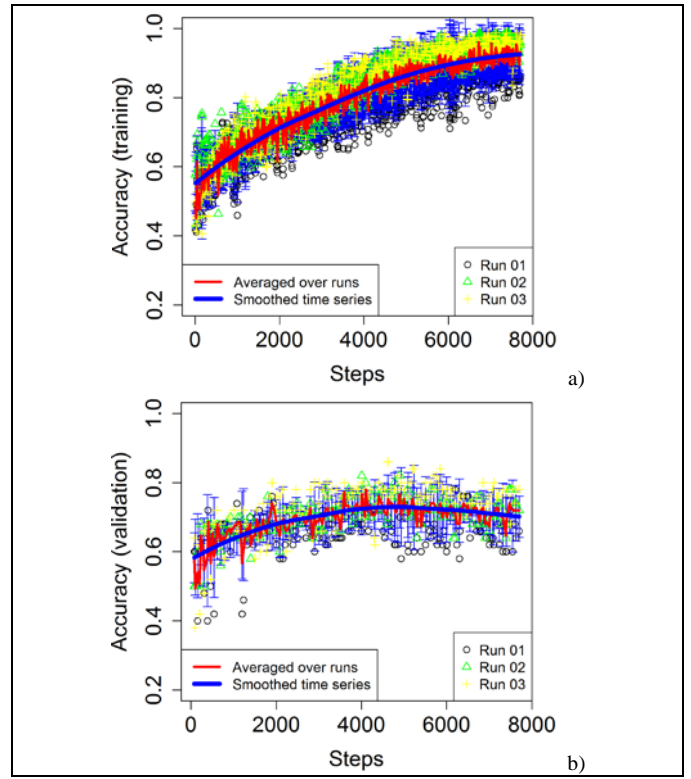

Fig. 7. Accuracy during some cross-validation runs for the segmented CXR images from SH dataset with lossless data augmentation: training (a) and validation (b).

But the most important aspect that validation loss (Fig. 8a) now has the global minimum (approximately at 3000 steps) that corresponds to the most realistic estimation of the validation accuracy, which is equal to 0.7±0.1.

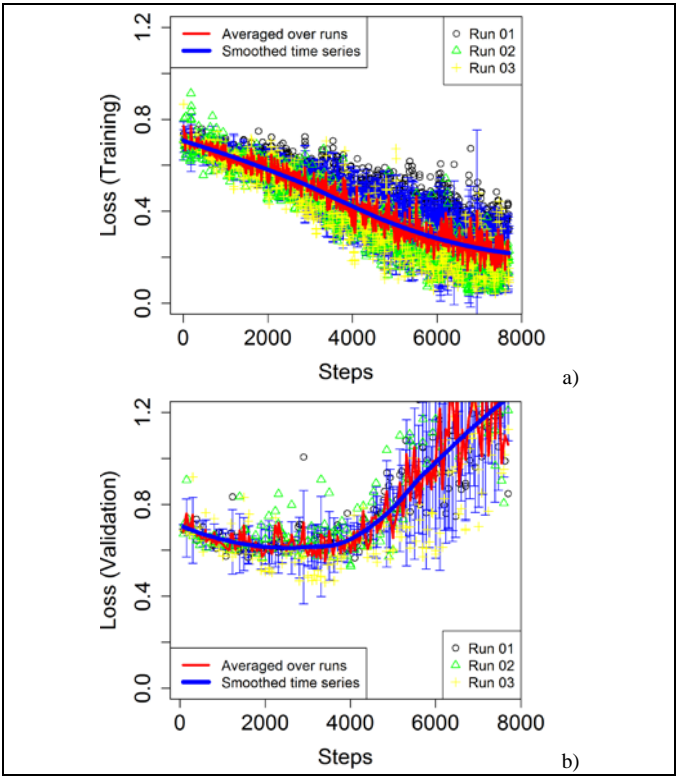

Fig. 8. Loss during some cross-validation runs for the segmented CXR images from SH dataset with lossless data augmentation: training (a) and validation (b).

## C. Effect of Lossy Data Augmentation

The additional lossy data augmentation for these 2D images included rotations by 5 degrees. In this case the training accuracy was lower than 0.85 (Fig. 9a) with the lower standard deviation (of the averaged and smoothed value), the validation accuracy was lower than 0.65 (Fig. 9b), and the training loss (Fig. 10a) was not lower than 0.35. But again the most important aspect that validation loss (Fig. 10a) has the global minimum (approximately at 2500 steps) that corresponds to the most realistic estimation of the validation accuracy, which is equal to 0.64±0.1.

Despite the lower number of training/validation runs for datasets with lossless data augmentation (3) and lossy data augmentation (2) in comparison to 7 runs for the segmented dataset, the standard deviation of the averaged and smoothed value (denoted by blue whisker lines in Fig. 5-10) of the validation accuracy was always lower for the augmented datasets (Fig. 7b and Fig. 9b). And the more interesting observation related with the much lower validation loss for the augmented datasets (Fig. 8b and Fig. 10b) in comparison to the segmented dataset (Fig. 6b), but just up to the moment when the validation loss reaches the local minimum (Fig. 8b and Fig. 10b), which corresponds to non-overfitted training.

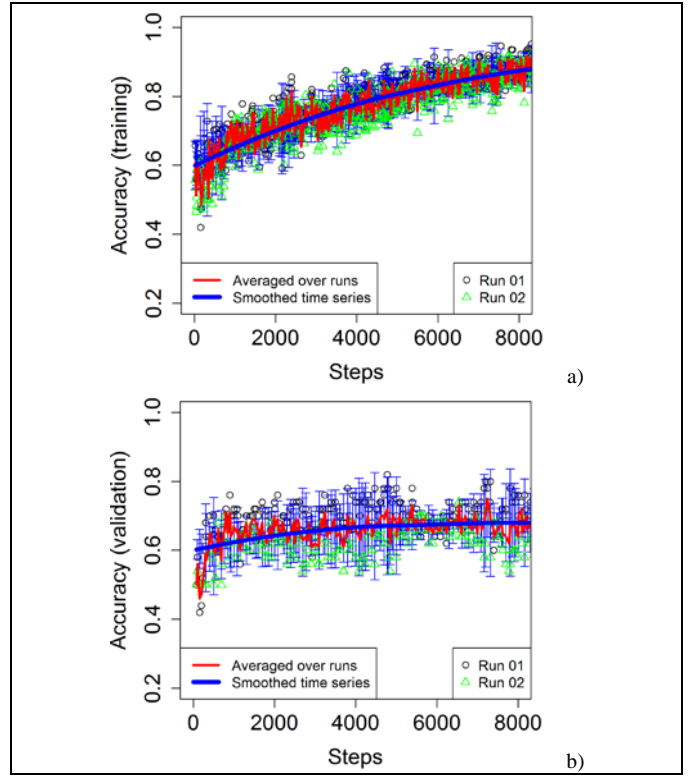

Fig. 9. Accuracy during some cross-validation runs for the segmented CXR images from SH dataset with additional lossy data augmentation: training (a) and validation (b).

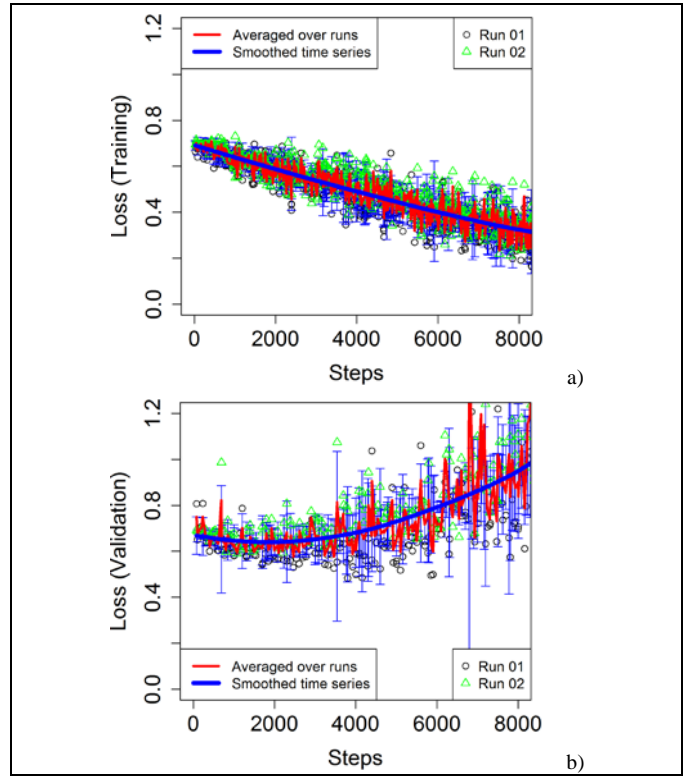

Fig. 10. Loss during some cross-validation runs for the segmented CXR images with additional lossy data augmentation from SH database: training (a) and validation (b).

## V. DISCUSSION AND CONCLUSIONS

The careful analysis of training and validation values of accuracy and loss (Fig. 11) for images from the segmented dataset and pre-processed datasets with lossless and lossy data augmentation allows us to derive the following observations and conclusions. The training rate is highest for the dataset obtained after lung segmentation (Fig. 11a, red lines), but it can be explained by the evident overtraining (overfitting). It is confirmed by the growth of the validation loss (Fig. 11b, red dash line) and the large difference between training (Fig. 11a, red solid line) and validation (Fig. 11a, red dash line) accuracies, and also by the large difference between training (Fig. 11b, red solid line) and validation (Fig. 11b, red dash line) losses. Despite the fact that the training rates for datasets with both lossless (Fig. 11a, blue solid line) and lossy (Fig. 11a, green solid line) data augmentations are much lower than the training rate for segmented dataset (Fig. 11a, red solid line), the data augmentation allowed us to avoid overfitting. Moreover, the loss values for both datasets with data augmentation are very close (Fig. 11b, green and blue lines) up to the moment when the validation loss reaches the local minimum (Fig. 8b and Fig. 10b), which corresponds to non-overfitted training. For the dataset with lossless data augmentation the training and validation accuracy values (Fig. 11a, blue lines) become equal at some stage, and validation accuracy reaches the maximum just near the local minimum of the validation loss.

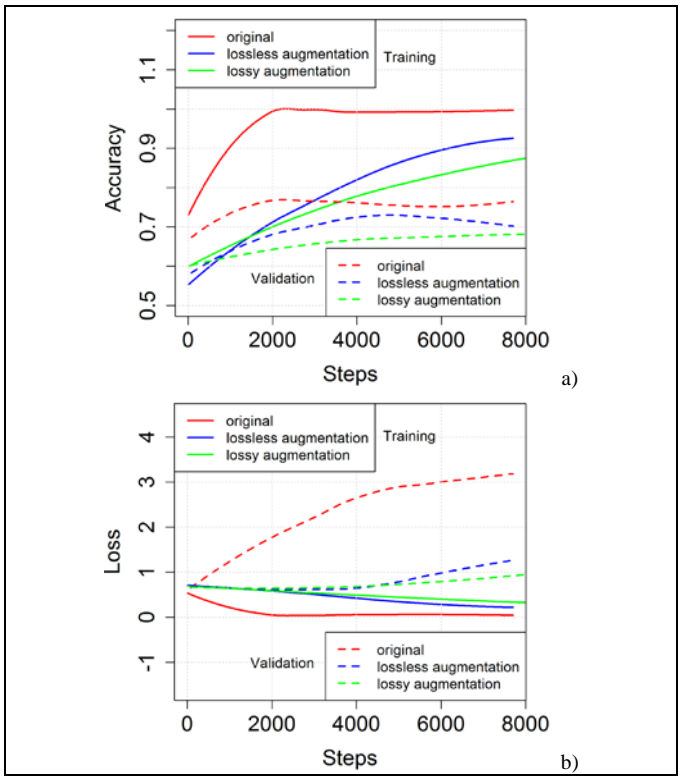

Fig. 11. The averaged and smoothed accuracy (a) and loss (b) values for the datasets noted in the legend: original segmented dataset (red color); the previous dataset with lossless data augmentation (blue); the previous dataset with additional lossy data augmentation (green).

But training and validation accuracy values for the dataset with lossy data augmentation never become close (Fig. 11a, green lines) and difference between them always grows during training and validation. This can be considered as a consequence of evident information loss during lossy data augmentation (actually rotation up to 5 degrees). Actually, it allows us to decrease the validation loss, but with decrease of validation accuracy. The different lossy data augmentation techniques can lead to the various positive and negative effects on CADx predictions [15], but the role of the information loss should be investigated more thoroughly. Moreover, influence of the outliers and long tails in lung area distribution can be excluded by naïve dimensionality reduction (due to "internal" bone segmentation) and t-SNE dimensionality reduction method with the significant increase of accuracy even for the very low portion of the filtered our outliers (5%) [6]. In this sense, the more careful exploratory data analysis and pre-processing with filtering out lung groups for people of various ages, genders, and geographical origin will be very promising.

The results obtained here demonstrate that the segmentation and data augmentation can improve the accuracy of prediction of lung diseases (availability of tuberculosis) even for the relatively small datasets like SH (662 images) [9,10]. Recently, the influences of inhomogeneity and outliers in CXR images was investigated for JSRT dataset with CXR images of cancer [5,6]. The pre-processed JSRT dataset obtained after lung segmentation, bone shadow exclusion, and filtering out the outliers by t-SNE demonstrated the highest training rate and best accuracy in comparison to the other pre-processed datasets. The current work on SH dataset with CXR images of tuberculosis shown that the similar lung segmentation technique allow to train (unfortunately with overfitting) deep CNN on such small and not well-balanced dataset like SH (that was impossible for the original SH dataset without segmentation). Moreover, the lossless data augmentation allows us to get the reliable training without overfitting and with steady validation loss decrease and global minimum. The effect of the additional lossy data augmentation (rotation by 5 degrees here) has the same tendency. But influence of lossy data augmentation in the wider sense (with more rotation, scaling, distortion, etc.) on CADx predictions for CXR images with marks of tuberculosis is open question yet and should be investigated along with other lung datasets [15].

The first evident way for improvement is related with increase of the datasets from the current number of 662 images (SH) up to > 100 000 images (ChestXRay dataset). In this sense the progress can be reached by sharing the similar datasets around the world in the spirit of open science, volunteer data collection, processing and computing [16-18]. And the second way is based on usage of more complex deep CNNs like CheXNet with >100 layers [4] and their fine tuning that can be very crucial for efficiency of the models used [19-21]. But in the context of this paper the better results in CADx-related research by deep learning could also be reached by:

- the more detailed exploratory data analysis with cutting the dataset in the better balanced subsets, for example, to separate lung images of children and adults (which have the very different lung area), to separate them by gender, nationality, age, substance abuse, professional activity, general health state, and other parameters of patients under investigations;

- the better "external" lung and "internal" bone segmentation [5];
- exclusion of non-evident outliers, e.g, by dimensionality reduction methods (like t-SNE or other) [6];
- lossless data augmentation by other means (shifts, translations, etc.);
- lossy data augmentation which under work now [15].

In conclusion, the results obtained here demonstrate the efficiency of lung segmentation, lossless and lossy data augmentation for CADx of tuberculosis by the deep CNN applied to the small and not well-balanced dataset even. The deep CNN demonstrates ability to train (despite overfitting) on the pre-processed dataset obtained after lung segmentation in contrast to the original not-segmented dataset. Addition of lossless data augmentation to the segmented dataset leads to the lowest validation loss (without overfitting) and nearly the same accuracy (within the limits of standard deviation) in comparison to the original and other pre-processed datasets after lossy data augmentation. The pre-processed dataset obtained after lung segmentation and the limited lossy data augmentation shows the lower validation loss with the lower validation accuracy. But the effect of lossy augmentation should be investigated more thoroughly with regard to the concrete datasets and their heterogeneity [15].

ACKNOWLEDGMENTS

The work was partially supported by Huizhou Science and Technology Bureau and Huizhou University (Huizhou, P.R.China) in the framework of Platform Construction for China-Ukraine Hi-Tech Park Project # 2014C050012001. The manually segmented lung masks were prepared by students and teachers of National Technical University of Ukraine "Igor Sikorsky Kyiv Polytechnic Institute" and can be used as an open science dataset under CC BY-NC-SA 4.0 license (https://www.kaggle.com/yoctoman/shcxr-lung-mask).

REFERENCES

[1] E. Smistad, T.L. Falch, M. Bozorgi, A.C. Elster, and F. Lindseth, "Medical image segmentation on GPUs – A comprehensive review," Medical image analysis, vol. 20(1), pp. 1-18, 2015.

[2] D. Shen, G. Wu, and H. Suk. "Deep learning in medical image analysis." Annual review of biomedical engineering, vol. 19, pp. 221-248, 2017.

[3] Y. LeCun, Y. Bengio, and G. Hinton, "Deep learning," Nature, vol. 521(7553), pp. 436-444, 2015.

[4] P. Rajpurkar, J. Irvin, K. Zhu, B. Yang, H. Mehta, T. Duan, D. Ding, A. Bagul, C. Langlotz, K. Shpanskaya, M.P. Lungren, and A.Y. Ng, "CheXNet: radiologist-level pneumonia detection on chest x-rays with deep learning," arXiv preprint arXiv:1711.05225, 2017.

[5] Yu. Gordienko, Peng Gang, Jiang Hui, Wei Zeng, Yu. Kochura, O. Alienin, O. Rokovyi, and S. Stirenko, "Deep learning with lung segmentation and bone shadow exclusion techniques for chest x-ray analysis of lung cancer," The First International Conference on Computer Science, Engineering and Education Applications (ICCSEEA2018), arXiv preprint arXiv: arXiv:1712.07632, 2017.

[6] Peng Gang, Jiang Hui, Wei Zeng, Yu. Gordienko, Yu. Kochura, O. Alienin, O. Rokovyi, and S. Stirenko, "Dimensionality Reduction in Deep Learning for Chest X-Ray Analysis of Lung Cancer," International Conference on Advanced Computational Intelligence (ICACI 2018), arXiv preprint arXiv:1801.06495 (2018).

[7] J. Shiraishi, S. Katsuragawa, J. Ikezoe, T. Matsumoto, T. Kobayashi, K. Komatsu, M. Matsui, H. Fujita, Y. Kodera, and K. Doi, "Development of a digital image database for chest radiographs with and without a lung nodule: receiver operating characteristic analysis of radiologists' detection of pulmonary nodules," American Journal of Roentgenology, vol. 174, pp. 71-74, 2000.

[8] S. G. Armato, et al., "The lung image database consortium (LIDC) and image database resource initiative (IDRI): a completed reference database of lung nodules on CT scans," Medical physics, vol. 38(2), pp. 915-931, 2011.

[9] S. Jaeger, S. Candemir, S. Antani, Y.X.J. Wáng, P.X. Lu, and G. Thoma, "Two public chest X-ray datasets for computer-aided screening of pulmonary diseases," Quantitative imaging in medicine and surgery, vol. 4(6), pp. 475-477, 2014.

[10] S. Jaeger, A. Karargyris, S. Candemir, L. Folio, J. Siegelman, F. Callaghan, Zhiyun Xue, K. Palaniappan, R.K. Singh, S. Antani, G. Thoma, Y.X.J. Wang, P.X. Lu, and C.J. McDonald, "Automatic tuberculosis screening using chest radiographs," IEEE transactions on medical imaging, vol.33(2), pp. 233-245, 2014.

[11] X. Wang, Y. Peng, L. Lu, Z. Lu, M. Bagheri, and R.M. Summers, "Chestx-ray8: hospital-scale chest x-ray database and benchmarks on weakly-supervised classification and localization of common thorax diseases," IEEE Conf. on Computer Vision and Pattern Recognition (CVPR), pp. 2097-2106, arXiv preprint arXiv:1705.02315, 2017.

[12] Yu. Gordienko et al., Deep learning by TPU-optimized models," (submitted).

[13] M. Abadi, et al., "TensorFlow: large-scale machine learning on heterogeneous distributed systems," arXiv preprint:1603.04467, 2016.

[14] W.S. Cleveland, E. Grosse, and W.M. Shyu, "Chapter 8. Local regression models," in Statistical Models in S, J.M. Chambers, T. Hastie, Eds. London: Chapman and Hall, 1997, pp.309-376.

[15] Yu. Kochura, Yu. Gordienko, S. Stirenko et al., "Data augmentation for semantic segmentation and deep learning for lung disease diagnostics," (submitted), 2018.

[16] N. Gordienko, O. Lodygensky, G. Fedak, and Yu. Gordienko, "Synergy of volunteer measurements and volunteer computing for effective data collecting, processing, simulating and analyzing on a worldwide scale," Proc. IEEE 38th Int. Convention on Inf. and Comm. Technology, Electronics and Microelectronics (MIPRO) pp. 193-198, 2015.

[17] N.N. Rather, C.O. Patel, and S.A. Khan, "Using deep learning towards biomedical knowledge discovery," IJMSC-International Journal of Mathematical Sciences and Computing, vol. 3(2), pp. 1, 2017.

[18] N. Gordienko, S. Stirenko, Yu. Kochura, A. Rojbi, O. Alienin, M. Novotarskiy, and Yu. Gordienko, "Deep learning for fatigue estimation on the basis of multimodal human-machine interactions," Proc. XXIX IUPAP Conference in Computational Physics, arXiv preprint arXiv:1801.06048, 2017.

[19] Yu. Kochura, S. Stirenko, A. Rojbi, O. Alienin, M. Novotarskiy, and Y. Gordienko, "Comparative analysis of open source frameworks for machine learning with use case in single-threaded and multi-threaded modes," XIIth Int. Scientific and Technical Conf. on Computer Sciences and Information Technologies (Lviv, Ukraine) arXiv preprint arXiv:1706.02248, 2017.

[20] Yu. Kochura, S. Stirenko, and Yu. Gordienko, "Comparative performance analysis of neural networks architectures on H2O platform for various activation functions," IEEE Int. Young Scientists Forum on Applied Physics and Engineering (Lviv, Ukraine), arXiv preprint arXiv:1707.04940, 2017.

[21] Yu. Kochura, S. Stirenko, O. Alienin, M. Novotarskiy, and Yu. Gordienko, "Performance analysis of open source machine learning frameworks for various parameters in single-threaded and multi-threaded modes," In Advances in Intelligent Systems and Computing II. CSIT 2017. Advances in Intelligent Systems and Computing, vol. 689, N.Shakhovska, V.Stepashko, Eds. Cham: Springer, 2017, pp. 243-256.